\documentclass{article}

 \usepackage[dblblindworkshop, final]{neurips_2025}

\usepackage[utf8]{inputenc} % allow utf-8 input
\usepackage[T1]{fontenc}    % use 8-bit T1 fonts
\usepackage{hyperref}       % hyperlinks
\usepackage{url}            % simple URL typesetting
\usepackage{booktabs}       % professional-quality tables
\usepackage{amsfonts}       % blackboard math symbols
\usepackage{nicefrac}       % compact symbols for 1/2, etc.
\usepackage{microtype}      % microtypography
\usepackage{xcolor}         % colors
\usepackage[frozencache,cachedir=minted-cache]{minted}
\usepackage{wrapfig}
\usepackage{footmisc}
\usepackage{graphicx}
\usepackage{amsmath}
\title{nanoTabPFN: A Lightweight and Educational Reimplementation of TabPFN }
\workshoptitle{nanoTabPFN: A Lightweight and Educational Reimplementation of TabPFN}

\author{Alexander~Pfefferle$^{* 1,2}$ \quad Johannes~Hog$^{* 2}$ \quad Lennart~Purucker$^{2}$ \quad Frank~Hutter$^{3,1,2}$ 
\\ 
$^1$ELLIS Institute Tübingen \quad
$^2$University of Freiburg \quad
$^3$Prior Labs \quad
\\
 $\texttt{pfeffera@cs.uni-freiburg.de}$\\
}

\begin{document}

\maketitle

\begingroup
\renewcommand{\thefootnote}{\fnsymbol{footnote}}
\footnotetext[1]{Equal contribution.}
\addtocounter{footnote}{-1}
\endgroup

\begin{abstract}
Tabular foundation models such as TabPFN have revolutionized predictive machine learning for tabular data.
At the same time, the driving factors of this revolution are hard to understand.
Existing open-source tabular foundation models are implemented in complicated pipelines boasting over $10\,000$ lines of code, lack architecture documentation or code quality.
In short, the implementations are hard to understand, not beginner-friendly, and complicated to adapt for new experiments.
We introduce nanoTabPFN, a simplified and lightweight implementation of the TabPFN v2 architecture and a corresponding training loop that uses pre-generated training data.
nanoTabPFN makes tabular foundation models more accessible to students and researchers alike. For example, restricted to a small data setting it achieves a performance comparable to traditional machine learning baselines within \textbf{one minute of pre-training on a single GPU} (160\,000$\times$ faster than TabPFN v2 pretraining).
This eliminated requirement of large computational resources makes pre-training tabular foundation models accessible for educational purposes.
Our code is available at \url{https://github.com/automl/nanoTabPFN}.
\end{abstract}

\section{Introduction}
\label{introduction}

The field of tabular data has recently been undergoing significant changes with the introduction of Tabular Foundation models. This revolution was started with the introduction of TabPFN~\citep{hollmann2023tabpfntransformersolvessmall} and continued with newer foundation models such as TabDPT~\citep{ma2024tabdptscalingtabularfoundation}, TabICL~\citep{qu2025tabicl}, LimiX~\citep{zhang2025limix} and TabPFN v2~\citep{hollmann2025accurate}. 
TabPFN v2 significantly improved over TabPFN by introducing a new architecture, a new prior for training data, and many small tricks in the inference pipeline.
This improvement comes with a significant increase in the complexity of its implementation, with the official repository currently boasting over $10\,000$ lines of Python code. Such complexity presents a substantial hurdle for researchers and students who want to understand, modify, or build upon TabPFN.

We solve this issue by introducing nanoTabPFN, a simplified and lightweight implementation of the  TabPFN v2 architecture and a training pipeline in under $500$ lines of code. 
We also provide an interface to load pre-generated training data.
Our code can be used to pre-train nanoTabPFN for small tabular prediction tasks within minutes.
The lightweight and modular design of our code allows users to quickly familiarize themselves with tabular foundation models and allows for fast iterations of research ideas for the prior, training pipeline, or architecture.
We believe that nanoTabPFN will serve as a first step in the journey of students towards TFMs and will make this field of research more accessible.

\textbf{The contributions of our work are:}
(1) our repository itself, containing a simplified reimplementation of the TabPFN v2 architecture, along with an in-depth explanation and (2) experiments showing that nanoTabPFN  achieves a performance comparable to traditional machine learning baselines within one minute of pre-training on a single GPU in a small data setting.

\section{Related Work}

nanoTabPFN is inspired by the success of minGPT~\citep{Karpathy2020minGPT} and nanoGPT~\citep{Karpathy2022nanoGPT}.
minGPT provides a minimal educational implementation of GPT-2~\citep{gpt2}, while nanoGPT advances this work with a performance-centric reimplementation.
nanoGPT is one of the most popular GPT-2 implementations, with over $47\,000$ stars on GitHub currently. 
It enabled many students and researchers to learn about large language models and bootstrap research, such as research on optimizers, including Muon \citep{jordan2024muon} or Adam-mini \citep{zhang2025adamminiusefewerlearning}.

\section{nanoTabPFN} \label{sec:method}

nanoTabPFN consists of two parts: the model architecture and its training loop.
In this section, we provide an in-depth explanation of both of these parts, as well as a small code example and a description of the differences to the original TabPFN v2 implementation.

\begin{figure*}[t]
\vskip 0.2in
\begin{center}
\centerline{\includegraphics[width=0.98\textwidth]{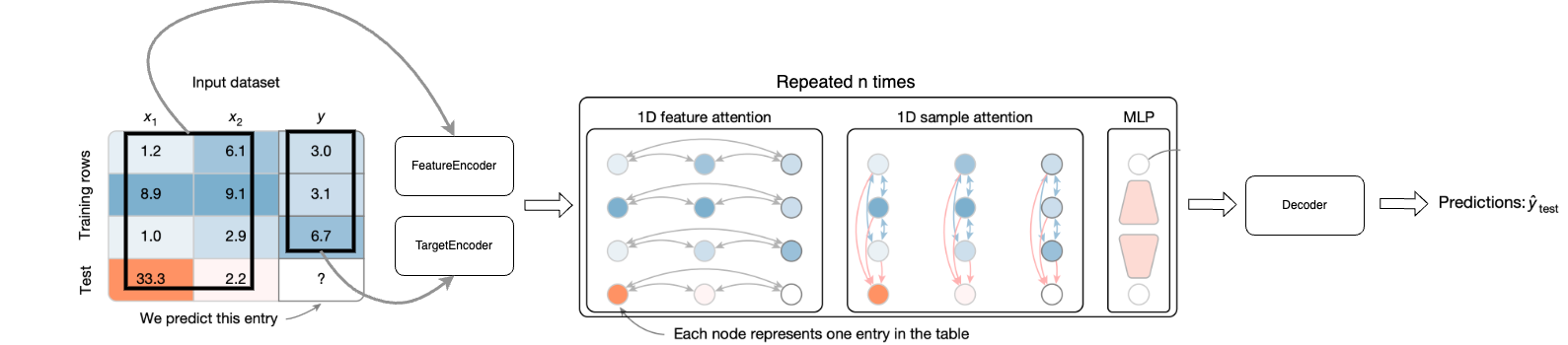}}
\caption{\textbf{nanoTabPFN Architecture.} The architecture consists of the \texttt{FeatureEncoder}, which normalizes and embeds the features, the \texttt{TargetEncoder}, which pads up the labels to the full length of rows and embeds the Targets, followed by a repeated \texttt{TransformerEncoderStack}, and the \texttt{Decoder}, which maps the high-dimensional embeddings to our predictions. Adapted from Figure 1 of \citet{hollmann2025accurate}.}
\label{overview-architecture}
\end{center}
\vskip -0.2in
\end{figure*}

\subsection{Model Architecture}
Figure~\ref{overview-architecture} illustrates the architecture of nanoTabPFN, which consists of four parts: 
the \verb|FeatureEncoder|, which normalizes the features and creates an embedding for each cell in the feature-part of the table;
the \verb|TargetEncoder|, which initializes the unknown test targets cells, and creates an embedding for each cell in the target-column of the table;
multiple \verb|TransformerEncoderLayers|, the main transformer layers adapting the embeddings of all cells in the table;
and the \verb|Decoder|, which maps the high-dimensional embedding of the unknown test targets to the predictions.

\begin{figure}[t]
\vskip 0.2in
\begin{center}
\centerline{\includegraphics[width=\columnwidth]{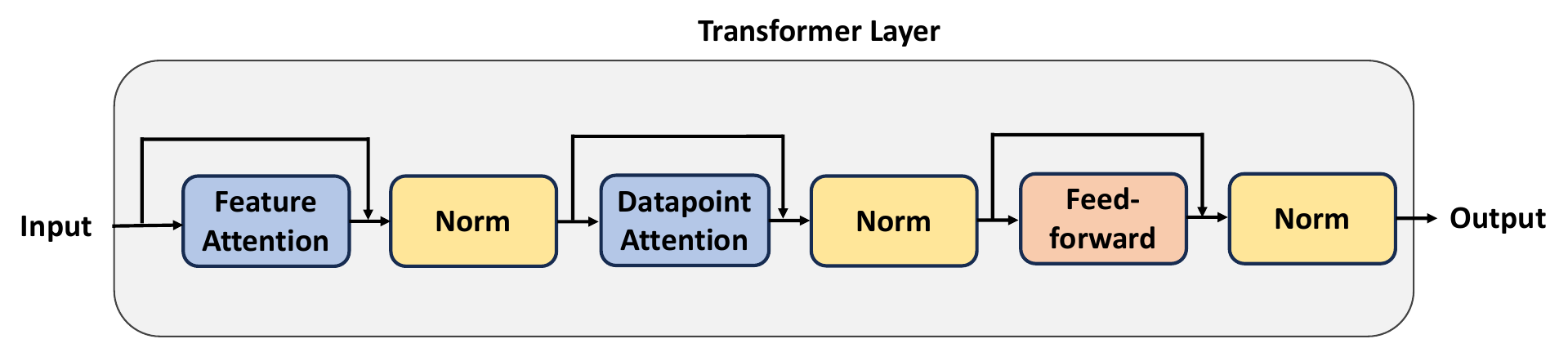}}
\caption{Transformer Layer. The Transformer Layer consists of Feature Attention, Datapoint Attention and a 2-layer MLP at the end. We have skip connections around each of the attention blocks and the MLP. A Layer Norm follows each skip connection.}
\label{transformer-layer}
\end{center}
\vskip -0.2in
\end{figure}

On an abstract level, TabPFN v2 works by alternating between applying attention between features and attention between datapoints on the table, as illustrated in Figure~\ref{transformer-layer}. 
To do this, we must first create an embedding for each cell in the table. 
We create the embeddings for all the features (\verb|X_train| and \verb|X_test|) using the \verb|FeatureEncoder|. 
The \verb|FeatureEncoder| normalizes each feature based on the mean and standard deviation of the training set, and removes outliers by clipping features that are too large or too small. 
Then it applies a linear layer to map the scalar values in the table to high-dimensional feature embeddings. 
Note that we use no positional embedding in our reimplementation since we want our model to be permutation-invariant with respect to datapoints and features.
The \verb|TargetEncoder| creates the embeddings for the target column; it extends \verb|y_train| with its mean values to create entries for \verb|y_test|, which we try to predict. Afterwards, it also applies a linear layer. 

Now that we have an embedding for each cell in the table we apply multiple \verb|TransformerEncoderLayers| sequentially, each of which applies bi-attention on the embeddings of the cells in our table (attention between features followed by attention between datapoints). 
In the attention between datapoints, the training data can only attend to itself and not the test data, while the test data can only attend to the training data.
This restriction ensures that the test data is not attended to, thereby guaranteeing that adding or removing datapoints to \verb|X_test| does not change the predictions for other datapoints. 

The \verb|TransformerEncoderLayer| has separate skip connections~\citep{he2016deep} around feature and datapoint attention, followed by layer normalizations \citep{ba2016layernormalization}.
After bi-attention the embeddings are further adapted by a cell-wise MLP, consisting of two feed-forward layers. 
This is surrounded by a skip connection and followed by a Layer Norm, see Figure~\ref{transformer-layer}.

The last part of our architecture is the \verb|Decoder|, which takes the embeddings of the \verb|y_test| cells as input and applies a 2-layer MLP. 
We treat the output of the MLP as logits for classification.

\begin{wrapfigure}{r}{0.50\textwidth}
\vspace{-1.4cm}
\begin{minipage}{0.50\textwidth}
\centering
\begin{minted}{python}
from model import NanoTabPFNModel
from train import PriorDumpDataLoader
from train import train

model = NanoTabPFNModel(
    embedding_size=96,
    num_attention_heads=4,
    mlp_hidden_size=192,
    num_layers=3,
    num_outputs=2
)

prior = PriorDumpDataLoader(
    "300k_150x5_2.h5",
    num_steps=2500,
    batch_size=32,
)

model, _ = train(
    model, 
    prior,
    lr=4e-3,
)
\end{minted}
\end{minipage}
\caption{Code example showing how to train nanoTabPFN.}
\label{fig:example-code}
\vspace{-1.8cm}
\end{wrapfigure}

\subsection{Training}

We provide a simple training loop that trains the model on the data given by a dataloader, using the scheduler-free~\citep{defazio2024road} version of the AdamW~\citep{adamw} optimizer without weight decay.
We support loading datasets that have been pre-generated and saved to the HDF5~\citep{The_HDF_Group_Hierarchical_Data_Format} format on disk. Enabling the loading of datasets from a file dump allows faster prototyping of the architecture and training code, since generating a new batch of data from a prior is quite expensive, and thus loading a pre-generated version drastically reduces the training time.

\subsection{Code Example}
On the right, we show a small example code for pretraining a 3-layer nanoTabPFN model on a pre-generated list of $80\,000$ datasets with exactly 150 datapoints, 5 features, and 2 classes. We later report results precisely for this code.

\subsection{Differences to TabPFN v2}
We intend nanoTabPFN to be an easier-to-understand and more lightweight version of TabPFN, and as such, we only include its core features. 
We do not include functionality of TabPFN that allows for combining neighboring pairs of features when creating the feature embedding, as it substantially increases code complexity, reduces interpretability and destroys permutation invariance of the features.
Another feature we do not include adds a column filled with row hashes to the table to differentiate between datapoints.
Finally, to keep our implementation minimal, we do not inherently handle categorical features and missing values.
If a user wants to evaluate nanoTabPFN on datasets containing categorical features or missing values, they have to preprocess them beforehand.

\section{Results}
\label{sec:results}
\begin{wrapfigure}{r}{0.5\textwidth}
\vspace{-2cm}
\begin{center}
\centerline{\includegraphics[width=0.5\textwidth]{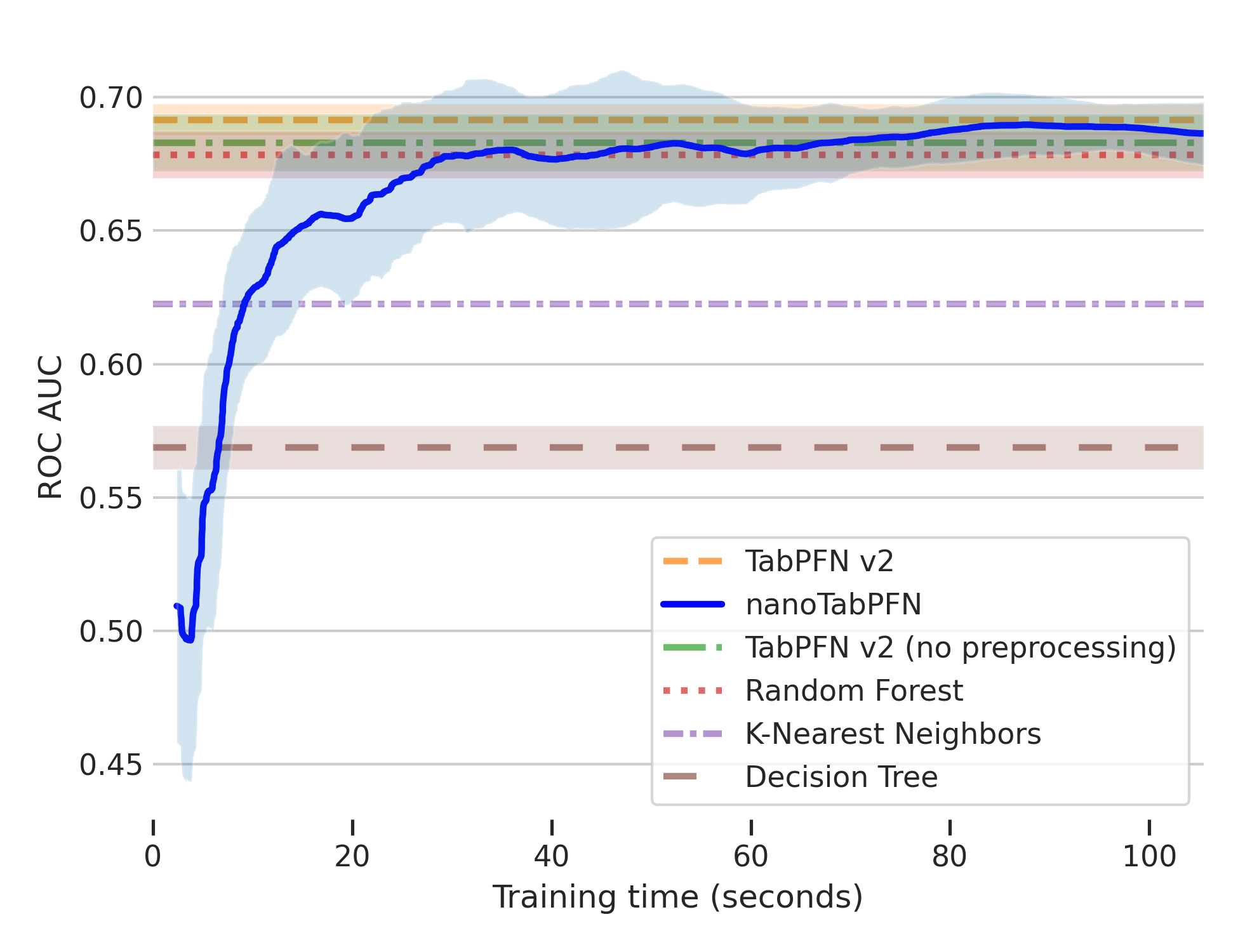}}
\caption{Within 60 seconds of pretraining on one consumer GPU, nanoTabPFN achieves average ROC AUC on a subset of subsampled datasets from TabArena comparable to traditional machine learning baselines.
}
\label{fig:runtime-plot}
\end{center}
\vspace{-0.95cm}
\end{wrapfigure}

We evaluate nanoTabPFN in a small data setting for educational use, demonstrating strong performance within minutes of pretraining.

We pretrained a small version of nanoTabPFN with 3 layers, 4 attention heads, an embedding size of 96, and a hidden layer size of 192 on $80\,000$ synthetically-generated datasets with exactly 150 datapoints and 5 features and 2 classes each, with a batch size of 32, using the example code from Figure~\ref{fig:example-code}.
Training converged after one minutes on a single NVIDIA GeForce RTX 2080 Ti GPU (11GB VRAM), whereas TabPFN v2 was pretrained on eight of these GPUs for two weeks. This is more than 160\,000$\times$ faster (14*24*60*8=161\,280), substantially lowering the barrier to entry.

The traditional baselines we use are k-nearest neighbors, a decision tree, and a random forest, in their default scikit-learn~\citep{scikit-learn} configuration.
For TabPFN we evaluate two configurations, one is the default configuration and the other disables ensembling and pre-processing.
The latter configuration more closely aligns with the feature set of nanoTabPFN, enabling a fairer comparison of the trained models rather than the surrounding pipelines. For more detailed information about our experimental setup and the evaluation strategy, including the datasets we used, we refer to Appendix~\ref{appendix:setup}.

Figure~\ref{fig:runtime-plot} shows nanoTabPFN's average ROC-AUC over its training time. 
Within 60 seconds of training, nanoTabPFN reaches a higher ROC AUC than all traditional machine learning baselines, thereby demonstrating its effectiveness. 
After only a few more seconds, the similar scale of prior and the evaluation setting allowed nanoTabPFN to quickly learn in this restricted setting, outperforming the more generally trained TabPFN configuration without preprocessing.
We present per-dataset results in Appendix \ref{appendix:per-dataset-results}.

\section{Conclusion}
\label{sec:conclusion}
In this paper, we introduced nanoTabPFN, a small and lightweight implementation of the TabPFN v2 architecture.
nanoTabPFN includes the core features of TabPFN, resulting in an implementation consisting of less than 500 lines of code, compared to the over 10\,000 lines of code of the TabPFN repository. 
This allows us to train a model within minutes that performs comparably to traditional machine learning algorithms on small datasets. 
nanoTabPFN's fast training speed, combined with our lightweight implementation, enables researchers and students to understand the inner workings of TabPFN more easily and enables faster prototyping of new research ideas, in the same way minGPT and later on nanoGPT did for the space of large language models.

Since nanoTabPFN focuses on simplicity and educational value, we did not include all features of TabPFN v2 which results in performance limitations. For example, we do not include code for regression, handling missing values and ensembling across different pre-processings of the data. Despite these limitations we are able to achieve good performance on small datasets. We intentionally focus on small-scale data, as the repository is aimed at educational value with small-scale compute.
Lastly, while nanoTabPFN's repository contains code for its architecture and training, it lacks a simplified prior implementation, which we leave to future work.

To conclude, with nanoTabPFN, we take an important first step towards democratizing the field of TFMs, lowering the barrier to entry, and accelerating research on TFMs. We look forward to nanoTabPFN being used in many university courses to teach TFMs.

\section*{Acknowledgement}
Funded by the European Union. Views and opinions expressed are however those of the author(s) only and do not necessarily reflect those of the European Union or the European Commission. Neither the European Union nor the European Commission can be held responsible for them. This work was supported by the European Union’s Horizon Europe research and innovation programme under grant agreement No 101214398 (ELLIOT).
\begin{center}
  \begin{minipage}{0.3\textwidth}
    \centering
    \includegraphics[width=\textwidth]{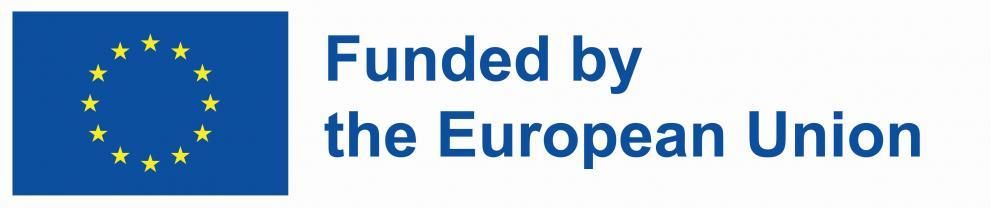}
  \end{minipage}
  \begin{minipage}{0.3\textwidth}
    \centering
    \includegraphics[width=\textwidth]{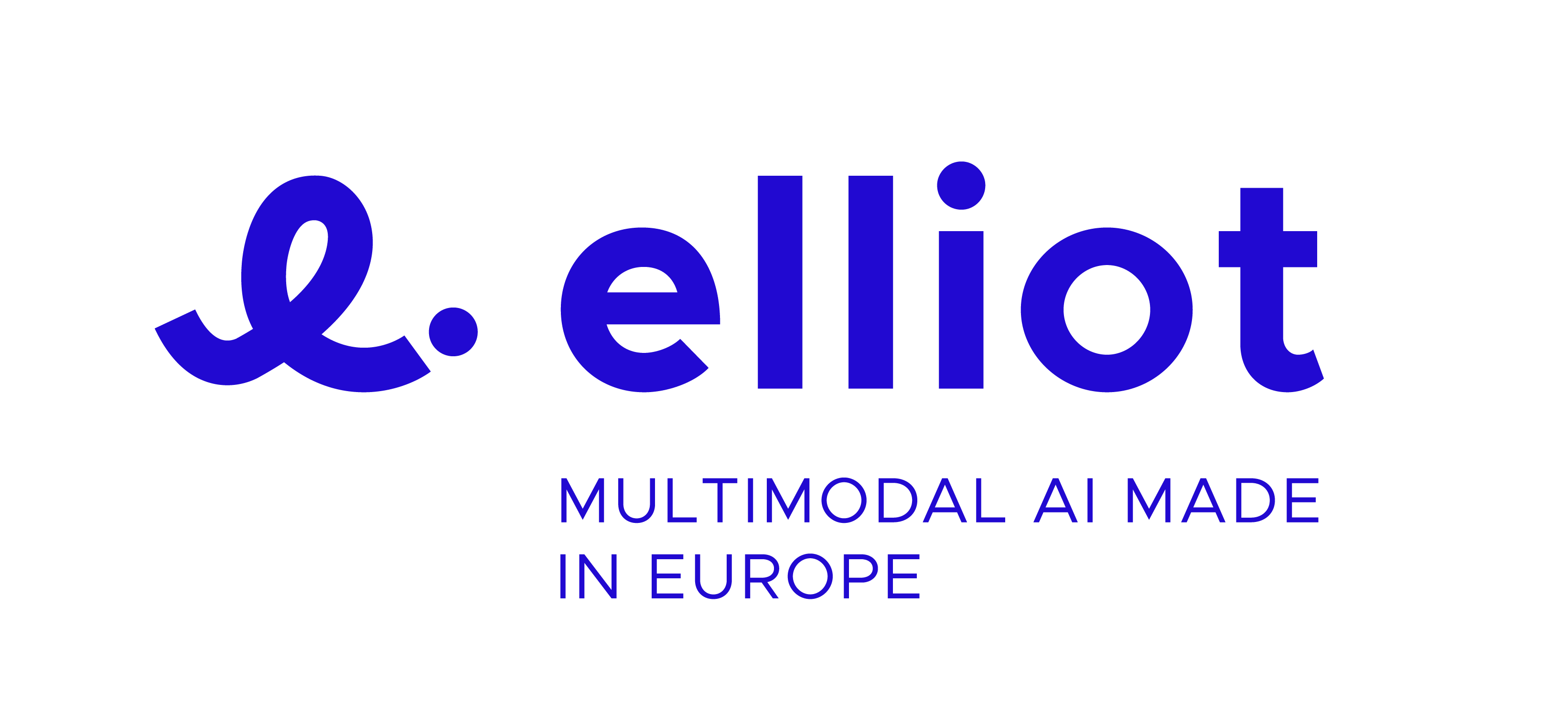}
  \end{minipage}
\end{center}
We thank the reviewers for their feedback, which contributed to improving this work. 
Additional we would also like to thank Jake~Robertson for his input regarding our prior selection.
Johannes~Hog and Lennart Purucker acknowledge funding by the Deutsche Forschungsgemeinschaft (DFG, German Research Foundation) under SFB 1597 (SmallData), grant number 499552394.
Frank Hutter acknowledges the financial support of the Hector Foundation.

\bibliographystyle{unsrtnat}
\bibliography{lib}

%%%%%%%%%%%%%%%%%%%%%%%%%%%%%%%%%%%%%%%%%%%%%%%%%%%%%%%%%%%%
\newpage
\appendix

\section{Detailed Experimental Setup}
\label{appendix:setup}

\paragraph{Hyperparameter Optimization}
We have chosen our prior, model, and training configuration based on a random search where we sampled 200 configurations.
The training time was limited to two minutes and we evaluated the performance on a synthetic prior consisting of 1600 datasets.
For generating training and evaluation data, we relied on TabICL's prior implementation.
The sampling was done on-the-fly and not included in the runtime measurement since we used pre-generated data when rerunning our best configuration. 
The search space and best configuration can be seen in Table~\ref{fig:search-space}.
For our comparison in Section~\ref{sec:results}, we rounded the configuration values before training.

\begin{table}[ht]
\centering
\caption{Hyperparameter Search Space and Optimal Configuration}
\label{fig:search-space}
\begin{tabular}{lll}
\toprule
\textbf{Hyperparameter} & \textbf{Search Space} & \textbf{Optimal Value} \\ \midrule
\texttt{lr} & $[10^{-4}, 5\times10^{-2}]$ (log scale) & 0.003892 \\
\texttt{weight\_decay} & $[10^{-9}, 10^{-4}]$ (log scale) & $1.00\times10^{-7}$ \\
\texttt{effective\_batch\_size} & \{8, 16, 32, 64\} & 32 \\
\texttt{num\_features} & [3, 13] & 4 \\
\texttt{num\_datapoints\_max} & [50, 300] & 154 \\
\texttt{num\_attention\_heads} & \{2, 4, 8\} & 4 \\
\texttt{embedding\_size} & \{64, 80, 96, 112, 128, 144, 160, 176, 192\} & 96 \\
\texttt{mlp\_multiple} & \{2, 4\} & 2 \\ 
\bottomrule
\end{tabular}
\end{table}

\paragraph{Experimental Setup}
We limited our evaluation to the binary classification datasets, which contain no missing values, in TabArena~\citep{erickson2025tabarenalivingbenchmarkmachine} with at most 10 features and subsampled the number of datapoints to 200.
The evaluation employed stratified 5-fold cross-validation with 20 repetitions. 
We only measured the accumulated training time and excluded the evaluation time at each step.
Our baselines used scikit-learn version 1.6.1 and tabpfn version 2.2.1.

\section{Additional Results}
\begin{figure}[h]
    \centering
    \includegraphics[width=\linewidth]{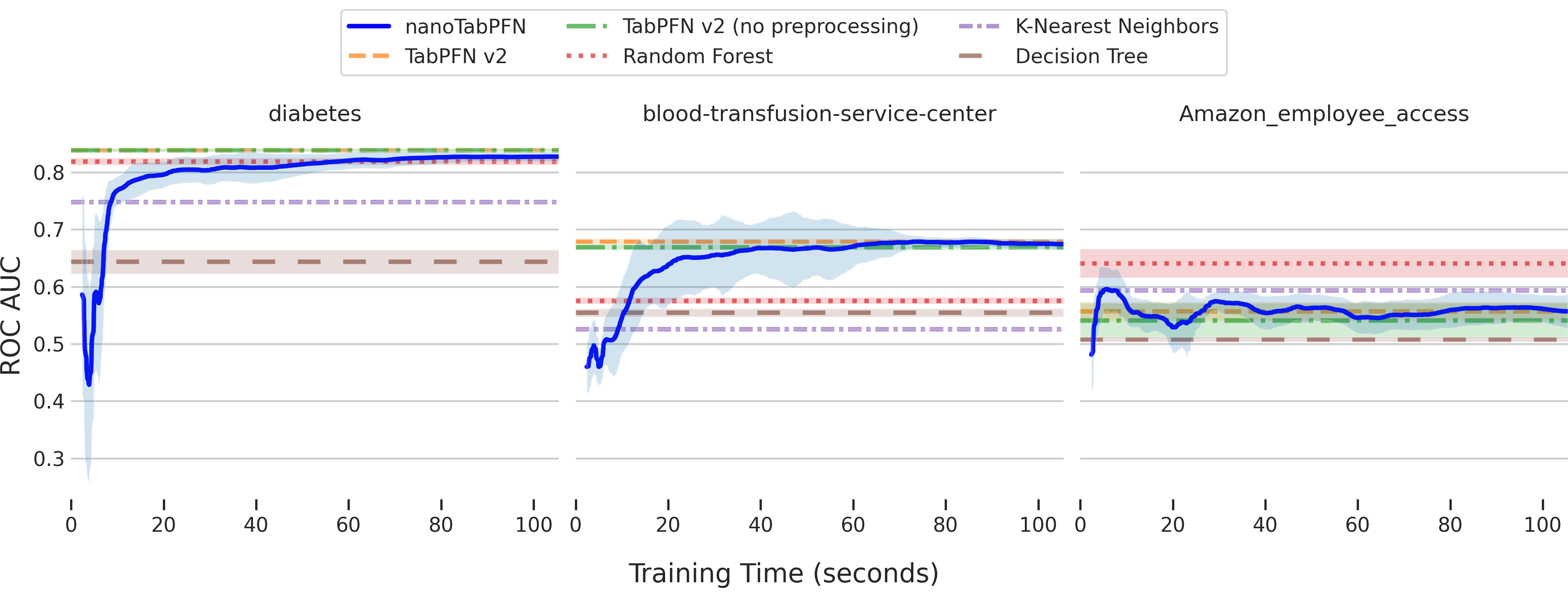}
    \caption{Per dataset results}
    \label{fig:individual-datasets}
\end{figure}

Figure~\ref{fig:individual-datasets} shows the ROC AUC during pretraining (Section \ref{sec:results}) on the individual evaluation datasets.
\label{appendix:per-dataset-results}
%%%%%%%%%%%%%%%%%%%%%%%%%%%%%%%%%%%%%%%%%%%%%%%%%%%%%%%%%%%%

\end{document}